\pdfoutput=1

\documentclass[11pt,a4paper]{article}

\usepackage[acceptedWithA]{tacl2018v2} 

\usepackage{times}
\usepackage{latexsym}

\usepackage[T1]{fontenc}

\usepackage[utf8]{inputenc}

\usepackage{microtype}

\usepackage{booktabs}
\usepackage{multirow}
\usepackage{graphicx}
\usepackage{xspace}
\usepackage{enumitem}
\usepackage{amsmath}
\usepackage{lscape}
\setlist{noitemsep}
\setlist[1]{labelindent=\parindent} 

\newcommand{\templama}{\textsc{Temp\-LAMA}\xspace}
\newcommand{\customnews}{\textsc{Custom\-News}\xspace}
\newcommand{\uniform}{\textit{Uniform}\xspace}
\newcommand{\temporal}{\textit{Temporal}\xspace}
\newcommand{\yearly}{\textit{Yearly}\xspace}

\newcommand{\blank}{\_\_X\_\_}

\title{Time-Aware Language Models as Temporal Knowledge Bases}

\author{Bhuwan Dhingra\thanks{\enskip Equal Contribution.} \qquad Jeremy R. Cole\footnotemark[1] \qquad Julian Martin Eisenschlos\\
        {\bf Daniel Gillick} \qquad {\bf Jacob Eisenstein} \qquad {\bf William W. Cohen}\\
        Google Research \\
        {\small \texttt{\{bdhingra,jrcole,eisenjulian,dgillick,jeisenstein,wcohen\}@google.com}}
        }

\begin{document}
\maketitle
\begin{abstract}
Many facts come with an expiration date, from the name of the President to the basketball team Lebron James plays for. However, most language models (LMs) are trained on snapshots of data collected at a specific moment in time. This can limit their utility, especially in the closed-book setting where the pretraining
corpus must contain the facts the model
should memorize.
We introduce a diagnostic dataset aimed at
probing LMs for factual knowledge that changes over time
and highlight problems with LMs at either end of
the spectrum---those trained on specific slices
of temporal data, as well as those trained on a wide range
of temporal data.
To mitigate these problems,
we propose a simple technique for jointly modeling text with its timestamp. This improves memorization of seen facts
from the training time period,
as well as calibration on predictions about unseen facts from future time
periods.
We also show that models trained with temporal context can be efficiently ``refreshed'' as
new data arrives, without the need for retraining from scratch.
\end{abstract}

\section{Introduction}
\label{sec:introduction}

Language models (LMs) have been
suggested as repositories of real-world 
knowledge~\citep{petroni2019language}
and there is much interest in using them
for tasks such as closed-book question answering~\citep[QA;][]{roberts-etal-2020-much},
fact verification~\citep{lee-etal-2020-language}
and dialogue~\citep{adiwardana2020towards}. 
Many facts, however, change with time. This raises two questions: Do pretrained LMs learn the appropriate temporal scope for the facts they encode? And what is the best way to update temporally-scoped knowledge in pretrained models?

Pretraining corpora for models such as BERT \cite{devlin-etal-2019-bert},
RoBERTa \cite{liu2019roberta} and GPT \cite{radford2019language}
are typically derived from a
snapshot of the web crawled at a specific moment in time \citep{raffel2019exploring}. While the impact on language modeling itself has been highlighted in recent work~\citep[e.g.,][]{lazaridou2021pitfalls,rottger2021temporal,hombaiah2021dynamic}, there are several potential problems specific to the encoding of factual knowledge:
\begin{itemize}
\item \textbf{Averaging}: For temporally-scoped knowledge, the model may see conflicting information, e.g., ``Lebron James plays for the Cavaliers / Lakers.'' Because LM training generally ignores temporal metadata, this can lead to an averaging effect, in which the model has low confidence in any of the correct answers.
\item \textbf{Forgetting}: Corpora such as Wikipedia and web crawls are constantly growing,
with documents distributed non-uniformly across time: there are more recent  documents than older ones, both because old documents can be updated and because more web documents are generated recently than in the past.
As a result, the model may fail to memorize facts that were valid only during underrepresented periods of time, and therefore do worse when asked questions about the more distant past.
\item \textbf{Poor temporal calibration}: As language models become ``stale'', they are increasingly likely to be queried about facts outside the temporal scope of their training data. While it may seem undesirable for a model to guess the answer to such questions, in many cases it is perfectly reasonable to assume that the future will be like the present: for example, in twenty years the capital of Alaska is unlikely to change, even though the \emph{governor} of Alaska is nearly impossible to predict. Ideally, the confidence with which the model responds to such queries should reflect this difficulty.
\end{itemize}

Temporally-scoped facts are common in practice;
however, QA datasets such as SQuAD \citep{rajpurkar-etal-2018-know}
or Natural Questions \citep{kwiatkowski-etal-2019-natural} focus on a single time period, even for questions whose answers are temporally scoped.
Thus, our first contribution in this paper is a diagnostic dataset, \templama
(short for TEMPoral LAnguage Model Analysis),
of
fill-in-the-blank queries for probing time-sensitive knowledge in LMs.
The queries in \templama are chosen such that the answer varies with time (\S~\ref{sec:templama}).
Using this dataset, we find empirical evidence of the problems mentioned above (\S~\ref{sec:experiments}).

As a first step towards addressing these problems, we propose a lightweight modification to pretraining.
We parametrize the masked language modeling objective~\citep[MLM;][]{devlin-etal-2019-bert} with temporal information, $P(y | x, t; \theta),$ where $y$ is a masked token or span, $x$ is the textual context, and $t$ is the time (\S~\ref{sec:methods-modeling}). 
The parameters $\theta$ must learn a representation of both text and time.
In the T5 framework~\citep{raffel2019exploring}, this can be accomplished by prefixing the input $x$ with a string representation of $t$,
e.g. ``year: 2018’’.
In addition, we pretrain from documents that are uniformly sampled from the timespan of the training corpus
which, in our case,
consists of news articles ranging from
$2010$-$2018$ \citep{lazaridou2021pitfalls} (\S~\ref{sec:methods-datasets}).
These interventions accomplish two goals: the model is exposed to facts from
the entire time range instead of just the most recent one,
which avoids forgetting certain temporally scoped facts.
Additionally, it prevents averaging because the facts are
assigned to different time buckets (in our case years).
This leads to improved recall of facts from the timespan
of the training corpus (\S~\ref{sec:exp-memorization}).

These interventions also improve the model's temporal calibration. 
We find that jointly modeling
text and time improves perplexity on future years unseen during training.
On \templama,
the joint model degrades more gracefully than a model unaware of time.
We also examine the model’s calibration farther into the future
using hand-crafted sets of queries whose answer is likely to
change \emph{frequently}, \emph{rarely}, or \emph{never}. 
We find qualitative evidence that the entropy of models
trained uniformly across the training timespan increases
most rapidly for the frequently-changing facts (\S~\ref{sec:exp-calibration}).

While calibration is desirable, models should be refreshed with new data when it becomes available. 
A standard practice for doing this is to combine the new and old data and retrain the model from scratch~\citep[e.g.,][]{liu2021newsembed}, but retraining can be costly for large-scale models~\cite{strubell-etal-2019-energy}.
On the other hand, finetuning only on the new data leads to catastrophic forgetting of the old data~\cite{zhu2020modifying},
since standard LMs have no knowledge of what is ``new'' and what is ``old'', unlike a model trained with temporal context.
We show that our temporally-scoped pretraining procedure makes LMs more amenable to post-hoc finetuning, as the data is implicitly bucketed into non-overlapping time slices.
We observe a similar performance to models retrained from scratch with $30 \times$ fewer steps,
and without degradation on the knowledge encoded by the older data (\S~\ref{sec:exp-adaptation}).

\paragraph{Summary of contributions:} (1) We offer \templama, a new dataset of temporally-scoped knowledge probes;
(2) We propose a simple modification to pretraining that facilitates the acquisition of temporal knowledge; (3) We conduct evaluations that demonstrate the impact of temporal shift on the knowledge encoded by existing LMs and the improvements offered by temporally-scoped pretraining; (4) We perform a qualitative analysis of temporal calibration into the future, again demonstrating the positive impact of temporally-scoped pretraining; (5) We show that temporally-scoped pretraining also facilitates efficient updates to existing pretrained LMs.

\begin{table}[!tb]
\setlength{\tabcolsep}{0.4em}
\centering
\footnotesize
\begin{tabular}{@{}clc@{}}
\toprule
\textbf{Year} & \textbf{Input}                                                                                                             & \textbf{Target}                                         \\ \midrule
\multicolumn{3}{c}{\customnews}                                                                                                                                                              \\ \midrule
2017          & \begin{tabular}[c]{@{}l@{}}The pound faces pressure from the US,\\ but the \_X\_ election could hit euro\end{tabular}      & French                                                  \\
2020          & \begin{tabular}[c]{@{}l@{}}\_X\_ accused Liverpool of 'crossing the\\ line' during win over his Chelsea side.\end{tabular} & \begin{tabular}[c]{@{}c@{}}Frank\\ Lampard\end{tabular} \\ \midrule
\multicolumn{3}{c}{\templama}                                                                                                                                                                \\ \midrule
2012          & Cristiano Ronaldo plays for \_X\_.                                                                                         & Real Madrid                                             \\
2019          & Cristiano Ronaldo plays for \_X\_.                                                                                         & Juventus FC                                             \\ \bottomrule
\end{tabular}
\caption{Examples from \customnews, which masks named entities and dates from news articles, and \templama, a novel synthethic dataset of temporally-scoped factual statements built from Wikidata.}
\label{tab:datasets}
\end{table}

\section{Methods}
\label{sec:methods}

\begin{figure*}[!htbp]
    \centering
    \includegraphics[width=\textwidth]{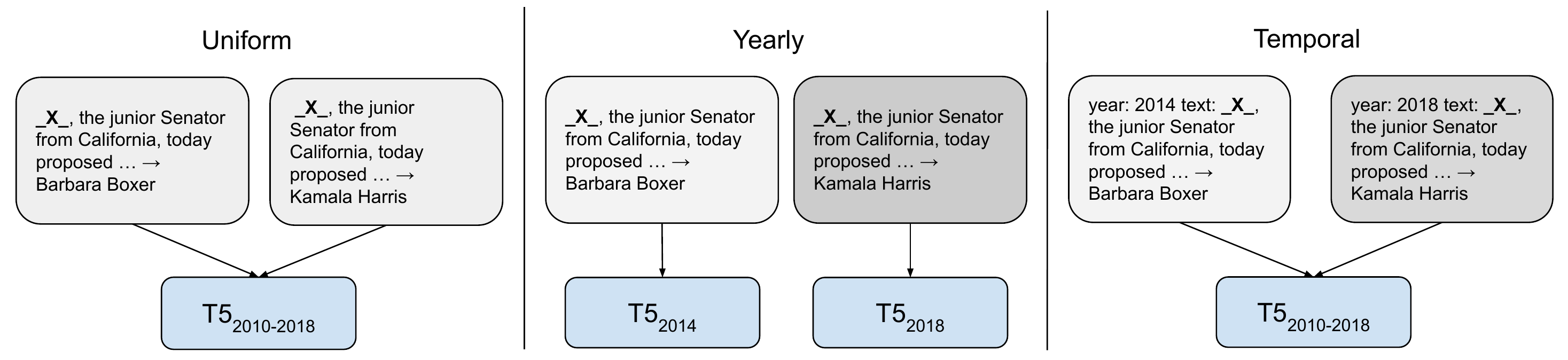}
    \caption{Three training setups to train T5 on \customnews: The \uniform model (\textbf{left}) is trained on all the data without explicit time information. The \yearly model (\textbf{middle}) avoids averaging over similar contexts by training separate models depending on the year, while the \temporal model (\textbf{right}) prepends a time prefix to each example.}
    \label{fig:data}
\end{figure*}

We probe factual knowledge in masked LMs using span prediction---given
an input statement $x$ with a span $y$ replaced by a special character, the task is to reconstruct that span.
Additionally, we assume that each $(x,y)$ pair has a timestamp $t$ denoting the
time at which it was written or a point in time at which its assertion is valid.
In this paper, we discretize $t$ into yearly buckets and leave more fine-grained
groupings (e.g. at the level of months or days) for future work.
For simplicity and efficiency, all of our models are text-to-text Transformers \citep{vaswani2017attention}
initialized from publicly available T5 checkpoints \citep{raffel2019exploring} and then adapted to more time-dependent datasets.
We first describe these datasets, followed by the approaches for jointly modeling
text and time.

\subsection{Datasets}
\label{sec:methods-datasets}
We experiment with a large-scale news corpus (\customnews) for pretraining our models,
combined with a smaller diagnostic dataset of factual queries (\templama) for
evaluation.

\paragraph{\customnews}
The \customnews dataset is a subset of web documents that are determined to be news~\cite{lazaridou2021pitfalls} and have an associated date
either extracted from the article's URL or from its html by looking
for a publication date.
We adapt this dataset in two main ways. 
First, we focus on a subset created by randomly sampling $1$M news articles from each of the years $2010$-$2020$ which had the maximum number of articles.
Second, while \citet{lazaridou2021pitfalls} used this data for classic autoregressive language modeling, we instead adapt it for the MLM objective. 
Specifically, we split the articles into sentences $x$ and then identify
\textit{salient spans} $y$ in the text corresponding to named entities and dates.
The salient span masking (SSM) paradigm
improves question
answering performance in both open-book \citep{guu2020realm}
and closed-book settings \citep{roberts-etal-2020-much}.
SSM restricts the inputs to those which have a higher chance of
requiring world knowledge and 
better aligns with our objective of measuring the factual knowledge captured by the LMs.
Following \citet{guu2020realm}, we identify named entities using a BERT-based tagger trained on CoNLL-2003 data \cite{tjong-kim-sang-de-meulder-2003-introduction}
and a regular expression for dates.

\paragraph{\templama}
\label{sec:templama}

We also construct a more targeted masked LM evaluation
for probing temporally sensitive
knowledge.
Starting with the November 2020 Wikidata snapshot \cite{vrandecic2014wikidata}
we first identify all facts which have either a start or an
end date after $2010$
and whose subjects and objects are both
entities with Wikipedia pages.\footnote{
We use SLING \cite{ringgaard2017sling} for preprocessing.}
Among these $482$K facts, we identify
subject and relation pairs which have multiple
objects at different times and select nine relations
with the most such subjects.
For these relations we manually write template
cloze queries (e.g. ``\texttt{Subject} works for \blank.'')
and populate them with the $1000$ most frequent subjects
per relation.
For each subject and each relation we gather all the objects
with their associated time interval and construct a separate
query for each year in that interval.
When intervals for the object entities overlap,
we add all of them to the list of correct answers.
The query and the corresponding year form the inputs
$x$ and $t$, while the object entity is the target $y$.
In total we construct $50,310$ queries across $11$
years.\footnote{
The \templama data is available at
\url{https://github.com/google-research/language/tree/master/language/templama}
}
Note that these type of cloze-style questions naturally follow
the salient span masking paradigm, where the answer to the question is the span to be masked.
\autoref{tab:datasets} shows examples from both \customnews~ and \templama.
A full list of the relations in \templama and their template queries is
included in \autoref{app:templama}.

\subsection{Training and evaluation}
We train and evaluate each of our models on a mixture
of \customnews and \templama. 
All models are initialized from a public T5 checkpoint, and then further \textit{adapted} for 300K steps on our data.
From \customnews we hold out $2000$ articles each for
validation and testing from each of the yearly subsets.
From \templama we reserve $10\%$ and $70\%$ of the queries from
each of the yearly subsets for validation and testing, respectively,
ensuring that none of the subject entities overlap between train, validation, or test sets.
Splitting along subject entities ensures that none of the facts required
to answer the test queries are seen during training on \templama~\cite{lewis-etal-2021-question}.
Instead they must be learned in an unsupervised manner either from the T5 pretraining
or when adapting to \customnews.
We train over the combination of the two training sets such
that for every $1000$ inputs from \customnews,
the model sees $1$ input from \templama.
Finetuning on a small disjoint set of queries
from \templama in this manner avoids issues due
to suboptimal prompts \cite{jiang-etal-2020-know,logan2021cutting}
by allowing the model to learn 
the expected format of queries and answers (e.g. ``Liverpool F.C.''
vs ``Liverpool'').

We also partition the data into two groups based on the year:
\textbf{2010-18} and \textbf{2019-20}.
Models are trained only on the former, but tested on both to
measure their performance for both seen and future time periods.
This split was informed by the fact that
the T5 checkpoints were
pretrained on web text extracted in April $2019$.
The main metric for evaluation is a token-level F1 score
between the predicted and ground truth targets,
computed in the same way as for the SQuAD benchmark \cite{rajpurkar-etal-2018-know}.
For \templama queries with multiple targets we take
the max F1.

\subsection{Jointly Modeling Text and Time}
\label{sec:methods-modeling}

Given a dataset of $(x,y,t)$ triples we model $P(y|x,t;\theta)$ using
variants of the T5 model where,
given $x$ as the input sequence, we
maximize the likelihood of the target sequence $y$.
We compare two approaches to condition the predictions on the time $t$ (also see \autoref{fig:data}).

\paragraph{\yearly}
In the first approach we use the temporal context by training
separate models specialized to different time buckets (in our case years),
so $P(y|x, t;\theta) = P(y|x; \theta_t)$.
Hence, we train an ensemble of nine T5 models adapted to each year
between $2010$-$2018$ for an additional $300$K steps.
When provided with a test input, this approach routes it to the appropriate yearly expert based on its timestamp.
If the timestamp falls outside $2010$-$18$, we
use the closest yearly expert (e.g. $2018$ for all test inputs $\geq 2018$).

\paragraph{\temporal}
Training a separate expert for each time slice reduces the averaging across conflicting contexts (\S~\ref{sec:introduction}),
but keeping an ensemble of large-scale LMs is undesirable in practice.
Moreover, there are regularities in how often facts change
(e.g. the FIFA World Cup happens every $4$ years, whereas NBA Championships happen every year), which a model specialized to a single time slice might not be able to learn.
Hence we also train a single T5 model on the entire dataset from $2010$-$2018$ for $300$K steps. In this model, the time $t$ is concatenated to the input,
i.e. $P(y|x,t;\theta) = P(y|t\oplus x; \theta)$, using a simple string representation of $t$ as a prefix for
the input $x$, e.g. ``year: $2014$''.  

\paragraph{Baselines}
The T5 checkpoints released by \citet{raffel2019exploring} are pretrained on long inputs with
multiple masks and cannot directly be tested using our factual knowledge probes.
Instead, we establish a baseline on the datasets introduced above using the pretrained models
from \citet{roberts-etal-2020-much},
which were
trained using SSM on Wikipedia for an additional $100$K steps.
This is referred to as \textit{T5-CBQA} (closed-book question answering).
We also experiment with additionally finetuning this model on \templama for $5$K steps (\textit{T5-CBQA-ft}).

To isolate the effect of time-aware pretraining, we also train a \uniform model,
which trains on the same uniformly sampled data as \temporal for the same number of steps, but without the time provided as an input.
During training, examples are shuffled rather than presented in chronological order.
Note that there are many ways of sampling training data across time,
and the optimal choice likely depends on the relative importance of memorizing
old versus recent facts.
Here we assume all time slices in the training data are equally important
and hence focus on uniform sampling.

\paragraph{Hyperparameters}
We primarily focus on the Large-sized T5 models with $770$M parameters, but we also investigate the scaling with size by comparing to the Small ($110$M) and XXL ($11$B) versions.
We use the same set of hyperparameters as \citet{raffel2019exploring},
with a batch size of $2048$, a fixed learning rate of $0.001$ and a dropout rate of $0.1$.
All our models are trained for a fixed number of $300$K steps,
except when adapting to new data (\S~\ref{sec:exp-adaptation}),
and then evaluated on the test set.
We found the loss on held out \customnews was still improving at the end of $300$K steps, but the overall trends were stable; to limit the experimentation time we did not explore longer training runs. 

\section{Experiments}
\label{sec:experiments}
We design several experiments to highlight the problems
around temporally-scoped knowledge in LMs and to test
whether they can be addressed by joint models of text and time.

\subsection{Memorizing Facts Across Time}
\label{sec:exp-memorization}
To understand the interplay of memorization and time, we examine the \templama and \customnews performance on the $2010$-$18$ slice. This permits us to analyze the forgetting and averaging effects discussed in \S~\ref{sec:introduction}
by comparing models trained on different slices of the data and with or without the temporal context.

\paragraph{Results}
\label{sec:memorization-results}

\begin{table*}[!tb]
\centering
\small
\begin{tabular}{lccccccc}
\toprule
\multirow{2}{*}{\textbf{Model}} & \multirow{2}{*}{\textbf{\#Parameters}} & \multicolumn{3}{c}{\textbf{CustomNews}}                & \multicolumn{3}{c}{\textbf{TempLAMA}}                  \\ \cmidrule(l){3-5} \cmidrule(l){6-8}
                                &                                        & \textbf{2010-18} & \textbf{2019-20} & \textbf{Overall} & \textbf{2010-18} & \textbf{2019-20} & \textbf{Overall} \\ \midrule
T5-CBQA                         & 737M                                   & 20.2             & 19.8             & 20.1             & 5.4              & 4.3              & 5.2              \\
T5-CBQA-ft                      & 737M                                   & 15.2             & 15.7             & 15.3             & 17.8             & 15.3             & 17.3             \\ \hline
Uniform                         & 737M                                   & 30.6             & 27.8             & 30.1             & 28.1             & 19.8             & 26.6             \\
Yearly                          & 6.6B                                   & \textbf{33.4}    & 26.7             & \textbf{32.2}    & 28.5             & 21.8             & 27.3             \\
Temporal                        & 737M                                   & 32.1             & \textbf{29.5}    & 31.6             & \textbf{29.6}    & \textbf{22.2}    & \textbf{28.2}    \\ \bottomrule
\end{tabular}
\caption{F1 scores of Large-sized model variants
for salient span mask prediction on \customnews and \templama.
\textit{T5-CBQA} is the pretrained model from \citet{roberts-etal-2020-much},
and \textit{T5-CBQA-ft} is further finetuned on \templama.
The \yearly model is an ensemble of $9$ models each finetuned on a 
yearly slice of the training data between $2010$ and $2018$.
We use the $2018$ model when testing on $2019$-$20$.
The \uniform and \temporal models are trained on the entire data from $2010$-$18$,
and the latter has additional temporal context.
The F1 scores are macro-averaged across the evaluation years. 
The \temporal model performs better on \templama,
which is focused only on temporally-scoped facts, as well as on the unseen years for \customnews.
}
\label{tab:main}
\end{table*}

\autoref{tab:main} shows performance
on the $2010$-$18$ test sets of \customnews and \templama.
\textit{T5-CBQA} and \textit{T5-CBQA-ft} fare significantly worse
on \templama ($17.8$) than the more standard Natural Questions benchmark
($28.5$, c.f. \citet{roberts-etal-2020-much}).
In particular, we find that training on the news domain
leads to significant improvements on the temporally scoped
knowledge required by \templama (comparing \textit{T5-CBQA-ft} and \uniform).
The two approaches which condition the predictions on time,
\yearly and \temporal, improve over \uniform which trains on the same data but without temporal context.
The \yearly ensemble, however, has linearly more parameters and requires linearly more compute to train.
For $2010$-$18$, the \yearly model performs better on \customnews,
which is far more likely to describe short-lived facts,
but the \temporal model is better on \templama,
where the facts typically span multiple years.
We further investigate the relationship between fact durations
and model performance below.

\begin{figure*}[!tb]
    \centering
    \includegraphics[width=0.32\textwidth]{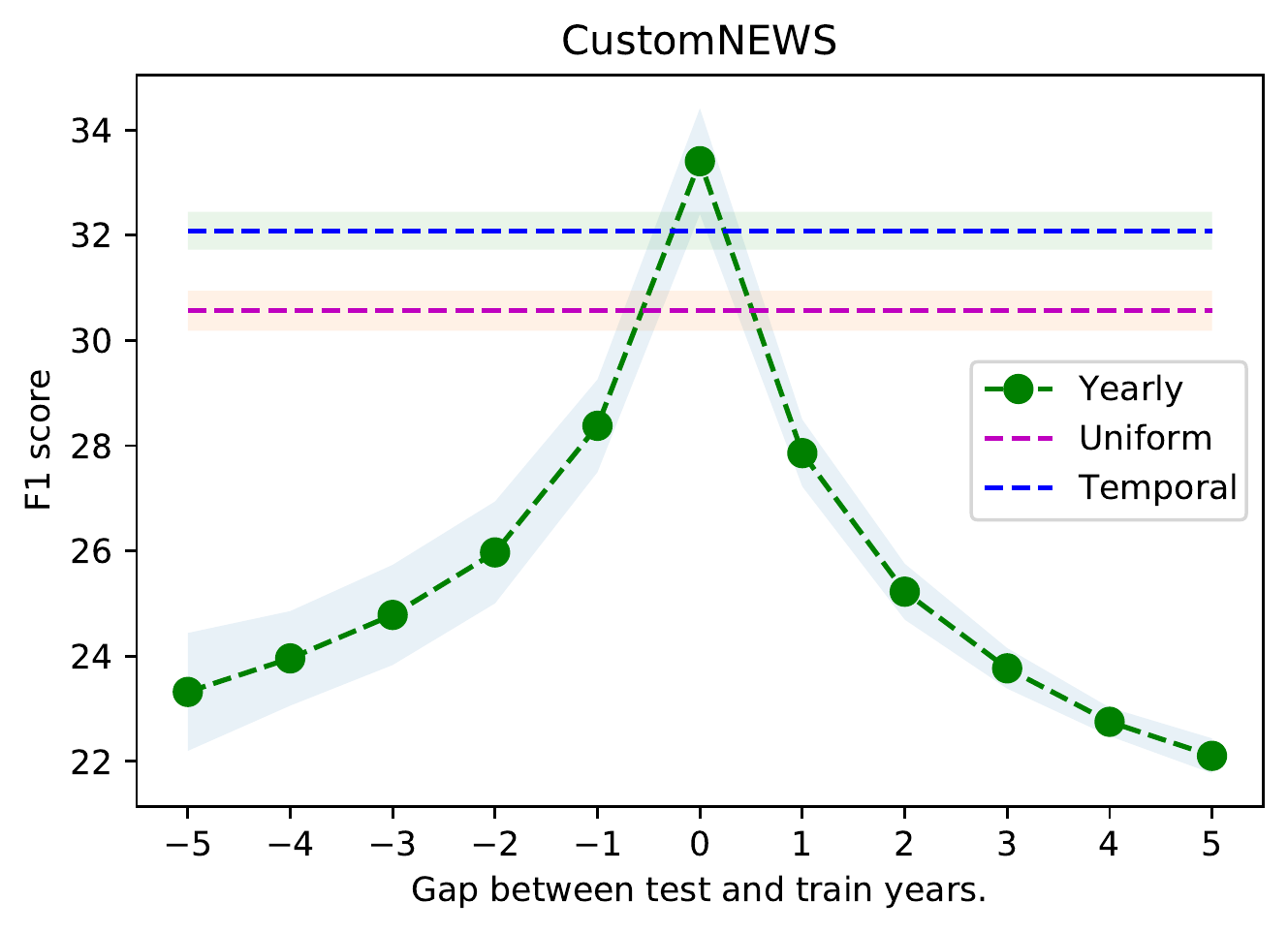}~
    \includegraphics[width=0.32\textwidth]{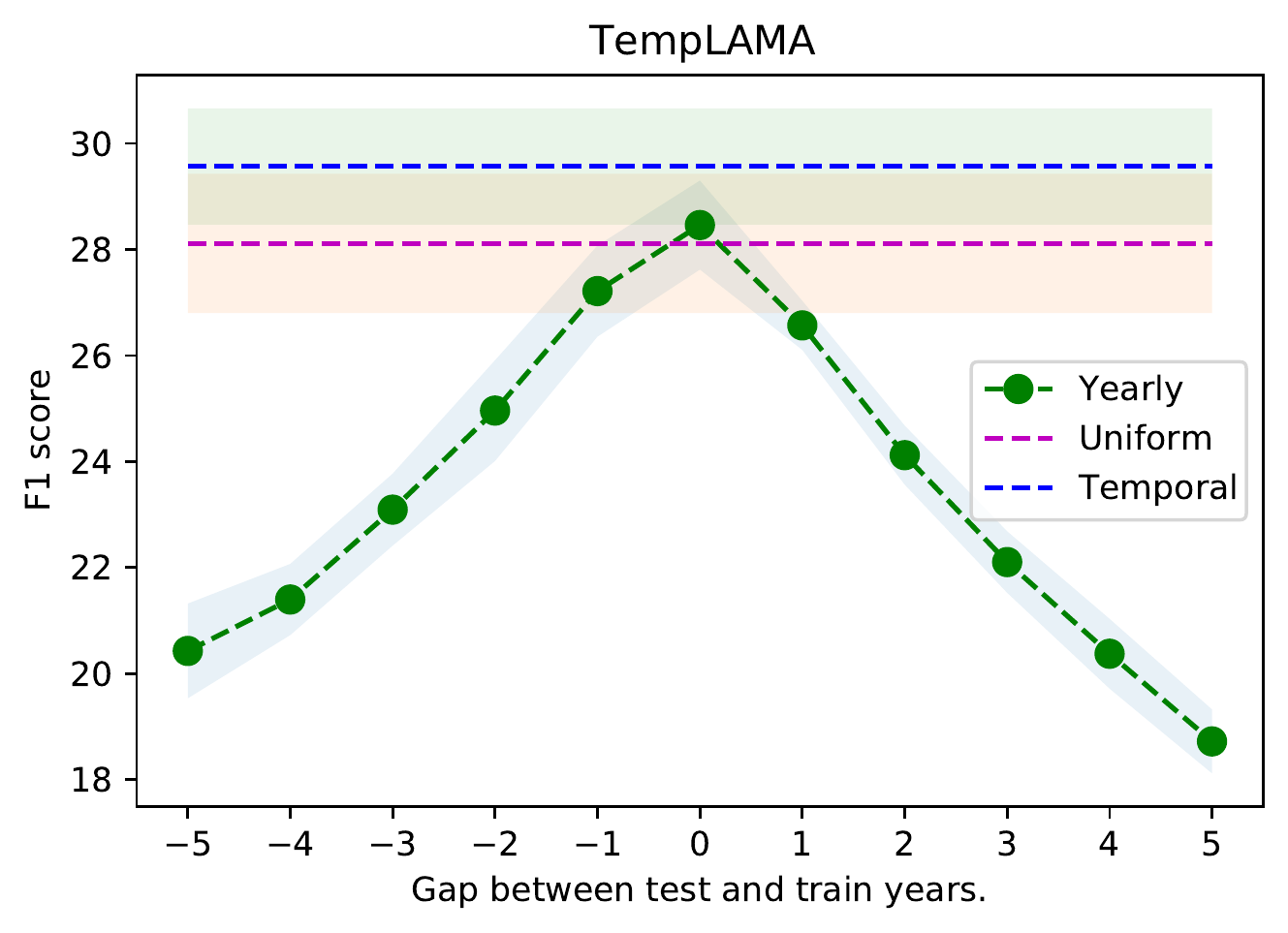}~
    \includegraphics[width=0.32\textwidth]{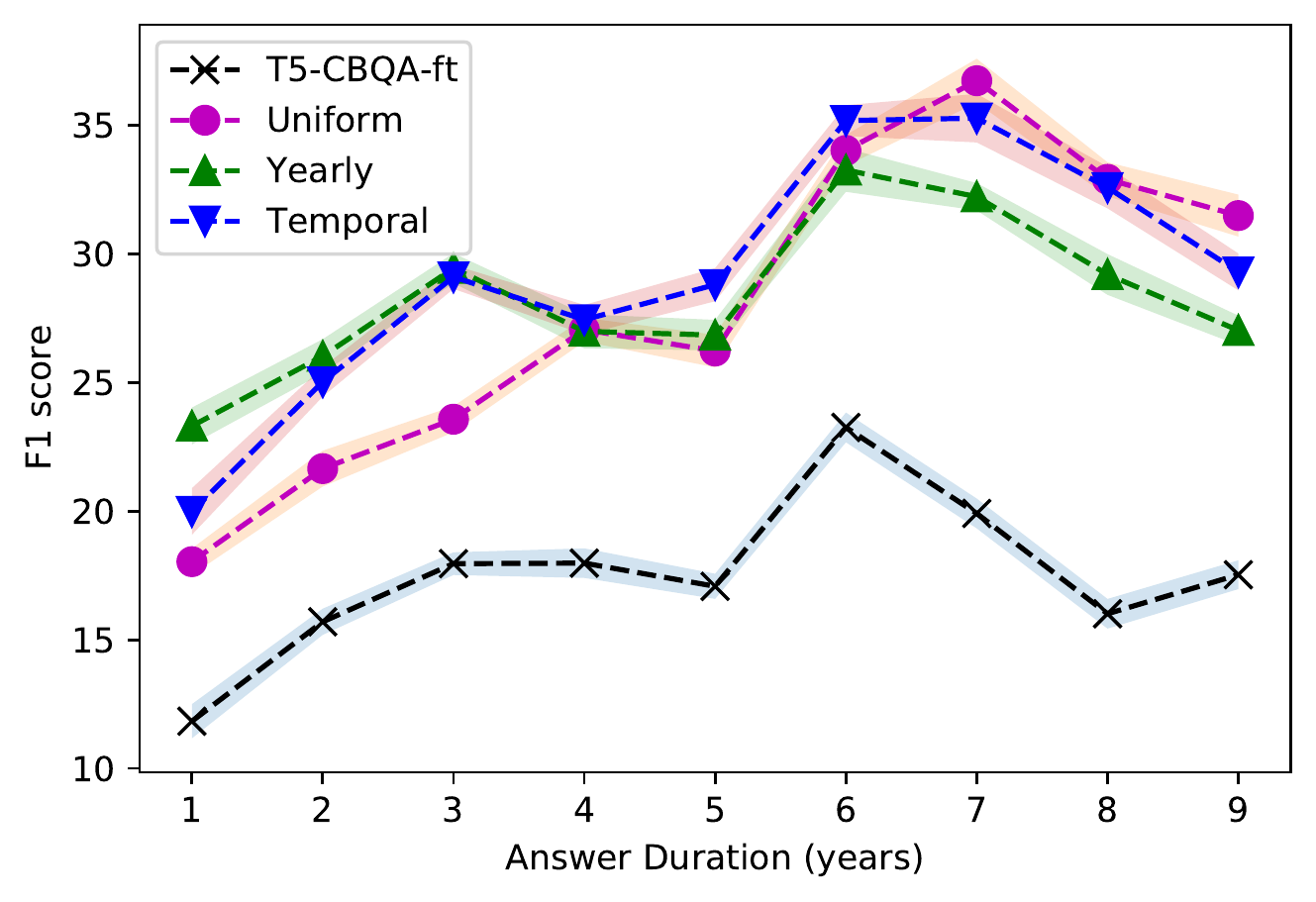}~
    \caption{F1 score of models trained on data from a specific year
    on \customnews (\textbf{Left}) and \templama
    (\textbf{Middle}) as the 
    gap between test and train years varies. Negative gaps
    indicate that the model is tested on data from \textit{before}
    the slice on which it was trained.
    The F1-score is macro-averaged across all possible pairs of train/test years between $2010$-$18$.
    For comparison we also show the F1 score of Uniform and Temporal models averaged across $2010$-$18$.
    Shaded area shows the $95\%$ confidence interval around the macro-average. The performance drop on both sides shows the forgetting effect.
    (\textbf{Right}) F1 scores on \templama
    grouped by the number of years for which 
    the answer to a query persists.
    Shaded area shows the $95\%$ confidence interval using bootstrap.
    }
    \label{fig:memorization}
\end{figure*}

We show empirical evidence of averaging and forgetting effects
in \autoref{fig:memorization},
which plots the F1 score of the year-specific models
as we vary the gap between test and train years.
The performance drops quickly on both sides, showing forgetting; however, the decline is larger for future years. 
The right plot compares F1-scores on \templama for queries grouped by the number of years for which their answer is valid.\footnote{
For multiple answers we pick the duration of the first one.
}
This is computed from the duration of their corresponding facts
in Wikidata.
The uniformly trained model has higher performance on queries whose answers persist for a long time, but it does worse on queries whose answers persist for less than $5$ years. The opposite is true for the year-specific models, which is intuitive due to the averaging effect of training on data from long periods of time.
Adding temporal context strikes a trade-off between these two extremes, leading to the overall higher F1 in \autoref{tab:main}.

\begin{table}[!h]
\centering
\small
\begin{tabular}{@{}ccccc@{}}
\toprule
\multirow{2}{*}{\textbf{Size}} & \multicolumn{2}{c}{\textbf{CustomNews}} & \multicolumn{2}{c}{\textbf{TempLAMA}} \\ \cmidrule(l){2-3}  \cmidrule(l){4-5}
                               & \textbf{Uniform}   & \textbf{Temporal}  & \textbf{Uniform}  & \textbf{Temporal} \\ \midrule
Small                          & 21.1               & 21.9               & 20.7              & 20.5              \\
Large                          & 30.1               & 31.6               & 26.6              & 28.2              \\
XXL                            & 32.3               & 33.8               & 28.4              & 30.5              \\ \bottomrule
\end{tabular}
\caption{Overall F1-score averaged from $2010$-$20$ for \uniform and \temporal models for different model sizes. Larger models benefit more from the temporal context.}
\label{tab:scaling}
\end{table}

\begin{table}[!t]
\small
\centering
\begin{tabular}{lr|cc}
\toprule
\textbf{Size} & \textbf{Model} & \textbf{EM} & \textbf{F1} \\
\midrule
\multirow{3}{*}{Small} 
& None & 3.63 & 9.51  \\
& Uniform & 4.01 & 10.27 \\
& Temporal & 4.05 & 10.20 \\ 
\midrule
\multirow{3}{*}{Large} 
&  None & 4.10 & 10.78 \\
& 2018 & 4.39 & 10.87  \\
&  Uniform & 4.70 & 11.34 \\
&  Temporal & 5.13 & 11.93 \\
\midrule
\multirow{3}{*}{XXL} 
&  None & 5.44 & 12.19 \\
& Uniform & 5.71 & 12.61 \\
& Temporal & 5.81 & 12.88 \\
\bottomrule
\end{tabular}
\caption{Test set results for models finetuned on the CronQuestions dataset in a closed-book manner.
``None'' refers to finetuning the T5 baseline; the ``$2018$'' model is adapted to the $2018$ slice of \customnews.
}
\vspace{-0.1in}
\label{tab:cronqa}
\end{table}

\begin{table*}[!tb]
\centering
\footnotesize
\scalebox{0.9}{
\begin{tabular}{llcc}
\toprule
\textbf{Input} & \textbf{Year}  & \textbf{Uniform} & \textbf{Temporal} \\
\midrule
\blank is the chair of Federal Reserve System. &	2019 &	Janet L. Yellen & Jerome Powell	\\
Nigel Farage is a member of the \blank. &	2019 &	UK Independence Party &	Brexit Party \\
Mark Sanford holds the position of \blank.	& 2017 & Governor of South Carolina & United States representative \\
\blank is the head of the government of New York City. & 2016 & Michael Bloomberg & Bill de Blasio \\
\blank is the head coach of Real Madrid CF. & 2015 & Zinedine Zidane & Carlo Ancelotti \\
Theresa May holds the position of \blank. & 2014 & Prime Minister of Great Britain & Home Secretary \\
Peyton Manning plays for \blank. & 2014 & Indianapolis Colts & Denver Broncos \\
\blank is the head of the government of United Kingdom. & 2011 & Theresa May & David Cameron \\
Marissa Mayer works for \blank. & 2011 & Yahoo & Google \\
Rahm Emanuel holds the position of \blank. & 2010 & Mayor of Chicago & White House Chief of Staff  \\ \bottomrule
\end{tabular}
}
\caption{Examples comparing the \uniform and \temporal models on \templama. The former frequently predicts a more common or newsworthy answer from the range of the training data, without taking the year into account.}
\label{tab:examples}
\end{table*}

Qualitatively, examining the \templama questions that the \temporal model answers correctly while the \uniform model answers incorrectly supports our hypothesis that the \uniform model is averaging over possible choices: it frequently answers with an entity that was more salient during our training period (see \autoref{tab:examples}).

\paragraph{Scaling}
\autoref{tab:scaling} shows the effect of increasing model size on the overall
F1 scores on \customnews and \templama.
In general, larger model sizes lead to a bigger improvement when training
with temporal context.

\begin{table}[!h]
\centering
\small
\begin{tabular}{@{}lccc@{}}
\toprule
\textbf{Model} & \textbf{2004-09} & \textbf{2010-18} & \textbf{2019-20} \\ \midrule
Uniform        & 34.8 ({\color{green}+6.3})      & 29.8 ({\color{red}-0.8})      & 27.4 ({\color{red}-0.4})      \\
Temporal       & 36.3 ({\color{green}+5.2})      & 31.1 ({\color{red}-1.0})      & 28.8 ({\color{red}-0.7})      \\ \bottomrule
\end{tabular}
\caption{F1 scores on different evaluation slices of \customnews
for models trained on data from $2004$-$18$.
Numbers in the parentheses show the absolute difference from
the same model trained on data from $2010$-$18$.}
\label{tab:timespan}
\end{table}

\paragraph{Longer Time Span.}
\autoref{tab:timespan} compares the Large-sized \uniform and \temporal models when
trained on a wider time period from $2004$ to $2018$.\footnote{
\customnews only has a small number of articles from $2003$ and before.
}
While the \temporal model still outperforms
\uniform, the gap is smaller between the
two compared to when training on $2010$-$18$.
In general increasing the time period
entails memorizing more facts for the
\temporal model.
Hence, this result suggests that the model
size should also be increased when training on
longer time spans.

\paragraph{CronQuestions}
To explore whether the improved memorization of facts
translates to downstream tasks,
we finetune the \uniform and \temporal models on
CronQuestions, a dataset of 410K time-dependent questions based on temporal knowledge graphs \citep{saxena-etal-2021-temporal}.
It consists of questions where the answer is either an entity or a temporal expression. Similar to \templama, the questions are based on Wikidata across time. 
We focus on a closed-book version of the task,
similar to the setup in \citet{roberts-etal-2020-much}, where the model is trained to predict the first answer in the list of correct answers for an input question.
During evaluation, it is compared to each answer in the set of
correct answers, and we take the maximum score among them.
\autoref{tab:cronqa} lists the SQuAD-based EM and F1 metrics on the test set.
We see an improvement in memorization for the \uniform and \temporal models, with the latter doing slightly better on the Large and XXL model sizes.

\subsection{Better Calibration in the Future}
\label{sec:exp-calibration}
We examine the model's performance on future slices of data at two different time scales. In the first, we look at \textit{graceful degradation}, mimicking the life-cycle of a model that has been deployed, and thus has not seen the newest slices of data yet. In the second, we ask the models to predict relations in the more distant future. While this may seem unreasonable, it is possible to articulate coherent intuitions about the future: for example, the capitals of U.S. states change far less frequently than their governors, and the probabilities emitted by language models should reflect this.

\subsubsection{Graceful Degradation}
Here we examine the \templama and \customnews performance on the $2019$-$20$ slices. Note that none of the models were pretrained or adapted to this slice, so these experiments allow us to measure degradation. 
We additionally look at the perplexity of the masked LM, which we compute as:
\begin{equation*}
    \textnormal{ppl} = \exp{-\frac{\sum_{(x,y,t)} \log P(y|x,t;\theta)}{\sum_y \textnormal{len}(y)}}.
\end{equation*}
Following \citet{lazaridou2021pitfalls}, we expect perplexity to increase for slices that are not covered in the training data, but we expect the temporally-conditioned model to be relatively more robust.

\begin{table}[!htbp]
\centering
\small
\begin{tabular}{@{}lcc@{}}
\toprule
\textbf{Model} & \textbf{2010-18} & \textbf{2019-20} \\ \midrule
T5-CBQA        & 26.11            & 29.22            \\
Uniform        & 11.68            & 14.37            \\
Yearly         & 13.62            & 23.30            \\
Temporal       & \textbf{11.33}   & \textbf{13.58}   \\ \bottomrule
\end{tabular}
\caption{Masked language modeling perplexity on \customnews (lower is better). The Temporal model degrades less when evaluated on the future time slice.}
\label{tab:news-perplexity}
\end{table}

\paragraph{Results}
Comparing the \textit{Uniform}
and \textit{Temporal} models in \autoref{tab:main},
we can see that
training with temporal context improves F1 scores on the $2019$-$20$ slices.
The \textit{Yearly} ensemble, which uses the latest $2018$ model when tested
on $2019$-$20$, is significantly worse on \customnews but comparable on \templama;
potentially because some of the answers remain the same.
A closer look at the model predictions
reveals that, unsurprisingly, none of the models are able to predict the \templama facts
that change after the training period.
Adding temporal context simply allows the \textit{Temporal} model to
persist the unchanged facts to $2019$-$20$.
On \customnews it has higher performance on the SSM objective, which includes both dates and entities in articles from an unseen time period.

\autoref{tab:news-perplexity} shows MLM perplexity on the \customnews test set.
The \temporal model has lowest perplexities on both the seen and unseen slices of evaluation data.
The \uniform model has lower perplexity than the \yearly one, especially on the future
slices where we use the $2018$ expert for the latter.
This suggests that, for language modeling, training
on more data outweighs the benefit of training on 
the specific temporal distribution of test data.

\begin{figure}[!htbp]
    \centering
    \includegraphics[width=\linewidth]{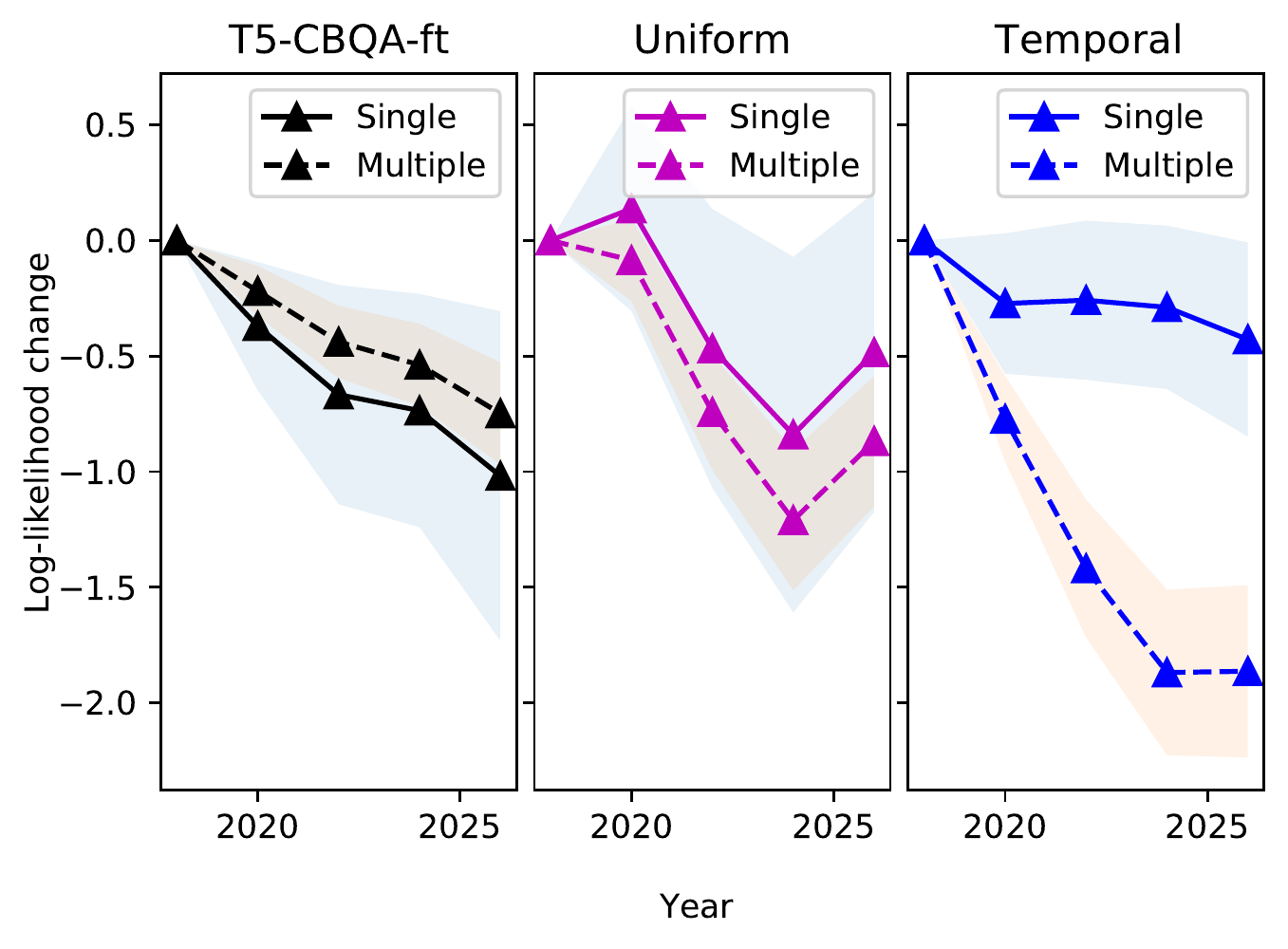}
    \caption{Change in log-likelihood over time of the most recent answer
    (from $2018$) 
    for \templama
    queries with \textit{Single} or \textit{Multiple} answers.
    The difference is taken from the value for
    the $2018$ answer. The \temporal model exhibits a more pronounced confidence gap for facts that changed in the past.
    }
    \label{fig:templama-confidence}
\end{figure}

Do the models learn how soon an answer is likely to change in the future?
We do a qualitative analysis by partitioning the \templama test queries
where each model was correct in the $2018$ evaluation into two
sets:
those with \textit{Single} or \textit{Multiple} answers across $2010$-$20$.
Then we measure the log-likelihood of that correct answer as we change the input year $t$ from $2019$ to $2029$, and plot the change in log-likelihood relative to $2018$ in \autoref{fig:templama-confidence}.
For the \textit{T5-CBQA-ft} and \uniform models, we vary the input
years by prefixing queries with ``In \texttt{year},...''.
The confidence for all models decreases as we get into the future,
which is reasonable since all relations in \templama are time-sensitive.
However, the confidence of the \temporal model
decreases
more rapidly for queries with multiple answers, reflecting the intuition that facts which have changed in the past are
likely to change again in the future.

\subsubsection{Future Relations}
\label{sec:future-relations} 
To further probe the models' understanding of expected versus unexpected
changes in the future,
we curate a small diagnostic dataset of queries about future relations.
We restrict the queries such that the answer is always either one of the $200$ largest US cities
or one of the $249$ countries in the world.
This allows us to compute the entropy of the predictions over a fixed set.
To relate model predictions to commonsense intuitions, we construct three sets of queries based on how frequently they are expected to change: \textit{frequent}, \textit{rare} and \textit{never}.
For example, the location of an awards show might change every year, while the city an athlete plays in changes every few years, and the location of a landmark almost never changes.
Then, given queries like ``In 2022, the Space Needle will be in \blank'' and ``In 2022, the NBA All-Star Game will be in \blank.'',
a model with a reasonable representation of time should have lower entropy for the former rather than the latter.
Moreover, the entropy should increase with time as the queries address the more distant future, and the \emph{rate} of increase should be greatest for frequently-changing relations. 
Note that we do not expect models to provide the \textit{correct}
answers for these queries (which we do not know anyway),
but only assign confidence in a manner consistent with human intuitions.
In total, we constructed $86$ queries across the three sets,
which are included in \autoref{app:future-relations}.

\begin{figure}[!t]
    \centering
    \includegraphics[width=\linewidth]{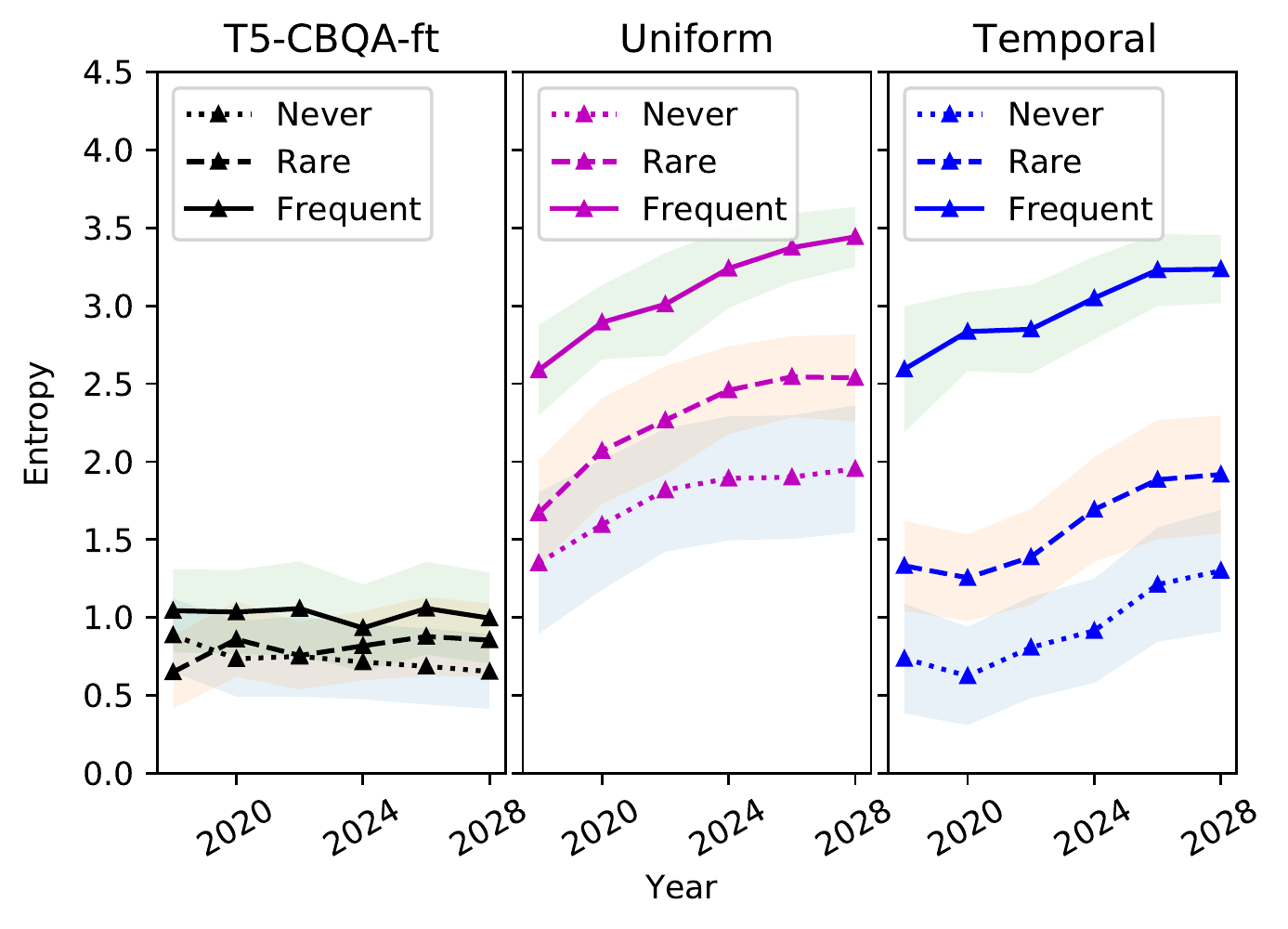}
    \caption{Entropy over time for \textit{frequent}, \textit{rare}, and \textit{never}-changing queries. The \emph{Temporal} model is more uncertain about frequently changing queries as time passes, and has a flatter entropy for constant facts.}
    \label{fig:generalization}
\end{figure}

\paragraph{Results}
\autoref{fig:generalization} shows the entropy of different
model variants averaged across the three sets of queries and plotted over time.
The baseline \textit{T5-CBQA-ft} model has a low constant entropy throughout,
irrespective of the query type.
Combined with its low accuracy on future slices from \autoref{tab:main},
this suggests it remains confidently incorrect and has poor calibration
about which facts are likely to change.
Both the \uniform and \temporal models have increasing uncertainty in the
future,
which is ordered correctly according to intuition: highest for
the queries of frequently-changing facts, and lowest for queries whose answers are expected not to change.
Interestingly, the \temporal model has a largely constant entropy
for rare- and never-changing queries until $2022$, after which it begins to increase.
While this agrees with intuition,
ideally a model should have low entropy on the never-changing set further into
the future. 


Overall, these results suggests that:
(1) models trained uniformly over a wide range of
time-sensitive data show improved calibration about expected changes
in the future;
and (2) training with temporal context further improves this calibration
for the first few years beyond the training period, in our case from
$2019$ to $2022$.
We also note the limitations with this evaluation, however:
(1) due to manual curation by the authors there are only $86$ queries in these sets,
and are likely to be biased in the facts they probe;
and (2) entropy mixes different kinds of uncertainty: that which is inherent
in the query (e.g. there are more distinct countries than cities with NFL teams),
as well as that due to the lack of confidence in the model.
We are interested in the latter, but our evaluation does not 
disentangle the two effects.


\begin{figure*}[!htbp]
    \centering
    \includegraphics[width=0.5\textwidth]{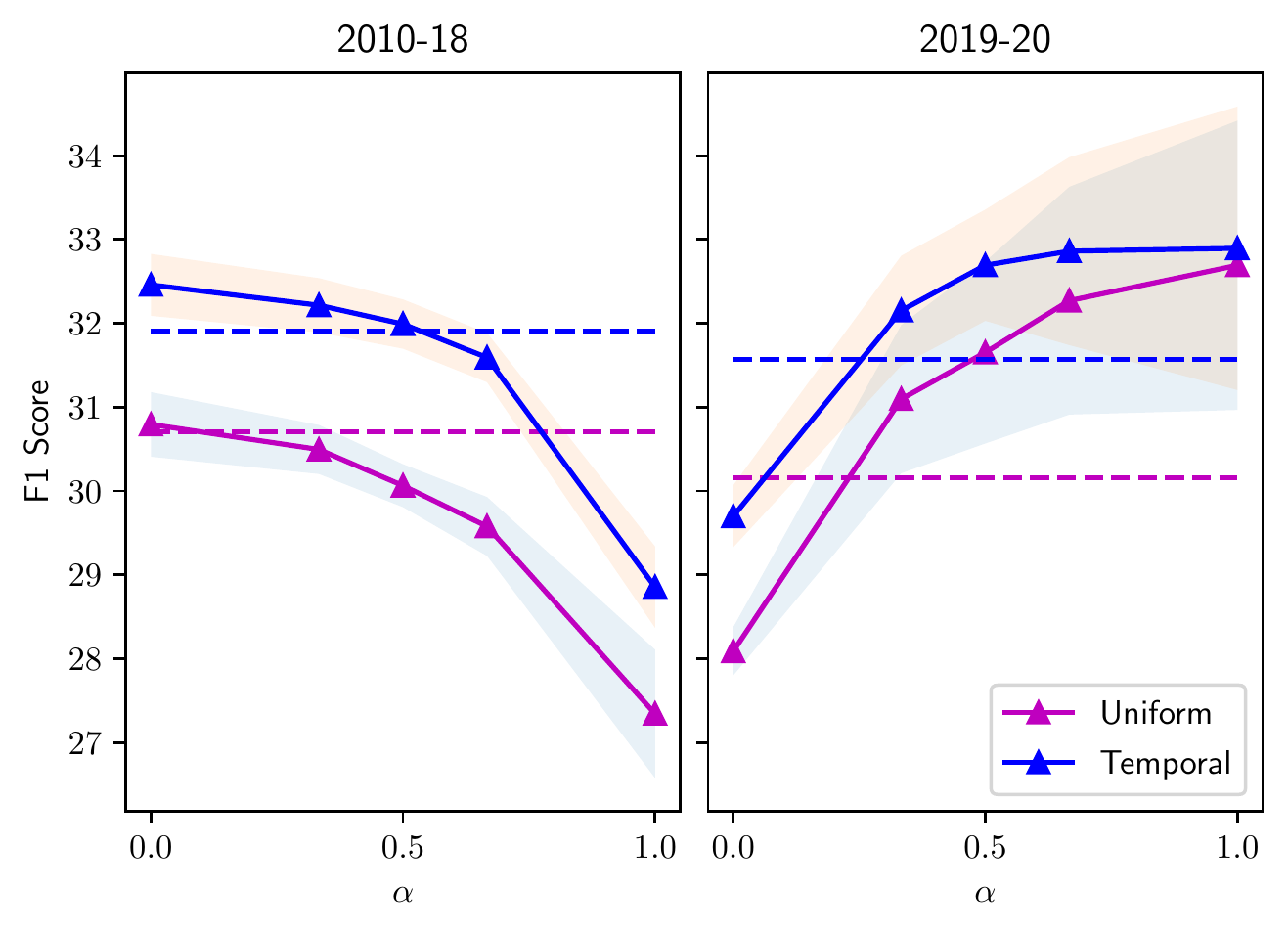}~
    \includegraphics[width=0.5\textwidth]{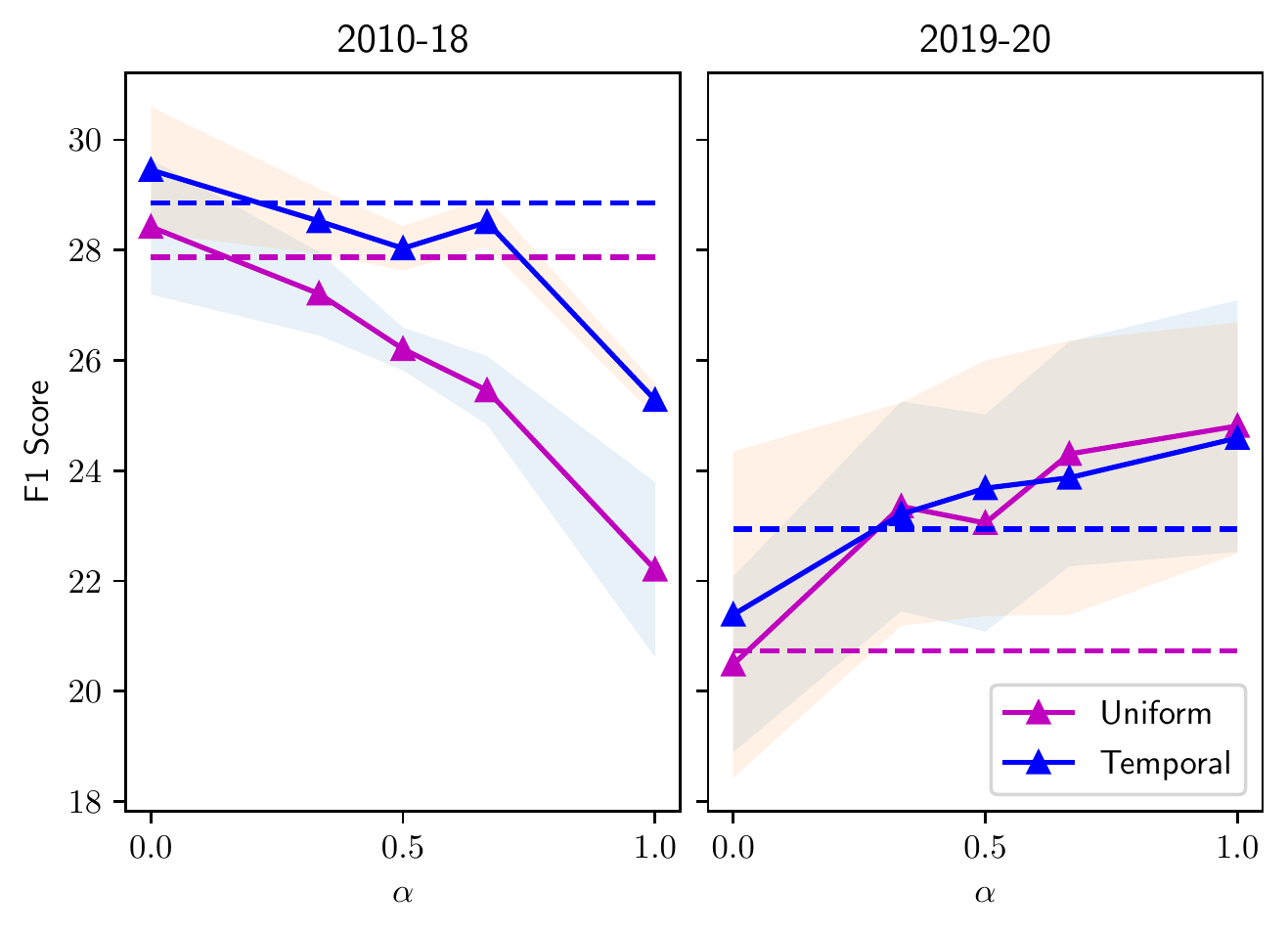}
    \caption{\customnews~(\textbf{left}) and \templama~(\textbf{right}) F1 score as models are \textit{adapted} to new data from $2019$ for $50$K steps.
    $\alpha$ denotes the fraction of training examples which
    come from the $2019$ slice (remaining examples come
    from the $2010$-$18$ slices).
    Dotted lines indicate models \textit{retrained} from scratch for $300$K steps on
    equal proportions of all data from $2010$-$19$. The \temporal model degrades less than \uniform on the $2010$-$18$ slice when adapted.}
    \label{fig:freshness}
\end{figure*}

\subsection{Cheaper Adaptation to New Data}
\label{sec:exp-adaptation}
Improved calibration about the future can help minimize mistakes
after the training time period (e.g. by abstaining),
but eventually models need to be refreshed as the world
changes and new data arrives.
In this section, we consider the setting where we have an already trained model
on the $2010$-$18$ slices,
as well as new data from the $2019$ slice.
We attempt to update the model on this new data
(as measured by the combined performance on $2019$-$20$
held out data) without forgetting the $2010$-$18$ slices.
These experiments are similar to the task posed by \citet{lazaridou-etal-2020-discovering}, but we compare the impact of adapting versus retraining from scratch. 
Finetuning only on the newest data ($2019$)
is suboptimal as the model forgets facts about the past (\autoref{fig:freshness}),
which was also observed by \citet{zhu2020modifying}.
Here we explore a simple alternative -- training on a mixture which
samples a data point from the new slice ($2019$) with probability $\alpha$
and a data point from the old slices ($2010$-$18$) with probability $1-\alpha$.
We finetune both the \temporal and \uniform models on this mixture for
an additional $50$K steps and compare the resulting performance
to models retrained from scratch for $300$K steps on data
sampled uniformly from all slices ($2010$-$19$).
Note that the latter strategy can be costly for large-scale LMs \cite{strubell-etal-2019-energy}.


\paragraph{Results}
\autoref{fig:freshness} shows the F1-score on \customnews~ and \templama~
as we vary $\alpha$. 
Across all values of $\alpha$, the \uniform model improves significantly on the $2019$ slice, but this comes at the cost of degrading on the $2010$-$18$ slices.
The \temporal model also adapts to $2019$,
but shows minimal degradation
on the $2010$-$18$ slice up to $\alpha=0.6$.
For $\alpha=0.5$ we found that
its performance with $10$K additional steps matches that of
the \temporal model trained from scratch for $300$K steps,
suggesting that models trained with temporal context can 
be efficiently adapted to new data without forgetting facts
from the old data.
\section{Discussion \& Limitations}
Our experiments have shown that current models have practical limitations in their ability to memorize the past and reasonably estimate the future.
These limitations can be mitigated by providing the model the date at which a text was created. While our results show consistent advantages, they also represent a narrow understanding of time.
In particular, the publication date of a news articles does not
necessarily correspond to the temporal scope of \textit{all}
events described in the article.
For example, articles may talk about historical events
or discuss events scheduled to happen in the future.
In \customnews around $3.9\%$ sentences explicitly mention a year between
$2010$-$18$, and $2.1\%$ mention the same year as the
publication date of the article.
This fraction is likely responsible for the improvement
of the \uniform model.
The \temporal model further assigns an approximate scope to
the remaining $96\%$ sentences and
it is encouraging to see improvements from that.
One avenue for future work is to explore
better strategies for assigning dates to these sentences.

We have focused on closed-book question answering, but temporal staleness of language models may have impacts in other applications as well. For example, in \emph{open}-book question answering, it is still necessary to align the question with relevant text in the retrieved passage, and this could be challenging when the question cannot be properly encoded by a stale LM: for example, the query ``which countries were affected by the 2020 hurricane season?'' would not match the passage ``Iota caused damages of \$564 million in Nicaragua'' in an LM that did not have access to training data mentioning ``Iota'' as a hurricane.

Another limitation of our work is that \templama is constructed in
a synthetic manner from WikiData. 
Incomplete or incorrect facts in the KB can result in
incorrect queries in \templama; for instance, we assume
a missing start date implies the fact is valid from the beginning of our time period of interest. We partition the \templama and \customnews dataset on the same yearly slices despite the nature of the datasets being quite different. Moreover, we did not investigate using longer or shorter temporal partitions.
Additionally, we did not test the ability to model temporal expressions such as ``before'' or ``during'', and we did not investigate temporal commonsense (e.g., \citealt{ZKNR19}), temporal ordering (e.g., \citealt{ning-etal-2020-torque}) or events (e.g., \citealt{zhou2020temporal}).

Lastly, it is worth noting that like all closed-book
models the models presented in this paper are also likely
to only memorize common facts about popular entities. This has the
danger of reinforcing stereotypes and leading to unfair outcomes.
Additionally, training the multitude of large-scale language models
presented in
this paper required the use of $32$ Cloud TPU v3 cores for several
hundred hours,
which has a significant environmental impact \cite{strubell-etal-2019-energy}.
However, our hope is that efficient schemes for updating temporally-sensitive
knowledge in LMs will eventually save energy costs in the long run.

\begin{table*}[htpb]
\small
\centering
\begin{tabular}{@{}clcl@{}}
\toprule
\textbf{WikiData ID} & \textbf{Relation}              & \textbf{\# Queries} & \textbf{Template}                                                                                         \\ \midrule
P54         & member of sports team & 9033       & \textless{}subject\textgreater~ plays for \textless{}object\textgreater{}.                        \\
P39         & position held         & 7343       & \textless{}subject\textgreater~ holds the position of \textless{}object\textgreater{}.            \\
P108        & employer              & 9049       & \textless{}subject\textgreater~ works for \textless{}object\textgreater{}.                        \\
P102        & political party       & 7324       & \textless{}subject\textgreater~ is a member of the \textless{}object\textgreater{}.               \\
P286        & head coach            & 4886       & \textless{}object\textgreater~ is the head coach of \textless{}subject\textgreater{}.             \\
P69         & educated at           & 1672       & \textless{}subject\textgreater~ attended \textless{}object\textgreater{}.                         \\
P488        & chairperson           & 4190       & \textless{}object\textgreater~ is the chair of \textless{}subject\textgreater{}.                  \\
P6          & head of government    & 4125       & \textless{}object\textgreater~ is the head of the government of \textless{}subject\textgreater{}. \\
P127        & owned by              & 2688       & \textless{}subject\textgreater~ is owned by \textless{}object\textgreater{}.                      \\ \bottomrule
\end{tabular}
\caption{Templates used for converting WikiData facts into natural language queries.}
\label{tab:templama-templates}
\end{table*}

\section{Related Work}

There is extensive prior work on learning diachronic embeddings of individual words~\cite[e.g.,][]{wijaya2011understanding,hamilton-etal-2016-diachronic,bamler2017dynamic}. Particularly related is the approach of \citet{dubossarsky-etal-2019-time}, who learn time-sensitive embeddings by concatenating each word token with the decade in which it appears.
As contextualized embedding models have largely replaced non-contextual word embeddings~\citep{peters-etal-2018-deep,devlin-etal-2019-bert}, the main application of diachronic word embeddings is to detect and model lexical semantic changes~\cite[e.g.,][]{frermann-lapata-2016-bayesian}, rather than to improve temporal awareness on downstream tasks. 
Our work fills this gap by adding a temporal component to T5, a pretrained language model that can complete multi-token spans. While \citet{giulianelli-etal-2020-analysing} use contextualized embeddings from BERT to model lexical semantic changes \emph{post hoc}, they do not add a time-sensitive component to the language model itself. Thus, their approach cannot support time-aware fact completion.

Several studies have focused on degradation of models
on test data from a different time period than their
training data
\cite{huang-paul-2018-examining,huang-paul-2019-neural,jaidka-etal-2018-diachronic,lukes-sogaard-2018-sentiment,komal2020time}.
\citet{delasalles2019learning} introduced an LSTM
language model which conditions
on dynamic author representations computed separately,
and showed that it improves perplexity
on both seen and unseen (future) time periods.
Most recently, \citet{rottger2021temporal} analyzed the interplay between
temporal adaptation during pretraining and finetuning,
and concluded that while both stages benefit from adaptation separately,
adaptation during pretraining does not help the downstream task.
Here we
show that the benefits of adaptation can be achieved using a single model that
conditions on time.
We further show that the benefits of adaptation come, at least in part,
from better memorization of time-sensitive facts.

In production contexts, an important form of temporal
generalization is the deployment of models trained on
data up to a certain time $T$ but applied on data after $T$: i.e., the present.
\citet{lazaridou2021pitfalls} show that language models gradually
degrade in performance under such a time-stratified
setting,
and propose \textit{dynamic evaluation} \cite{pmlr-v80-krause18a}
as a potential mitigation.
However, LMs are frequently applied to past data as well, e.g. for extracting
representations,
and here we show that updating on only the new data degrades performance on old data.
Our approach of conditioning on the temporal context alleviates this issue.

A related line of work has explored editing neural predictions
after training given a dataset of revised input and output pairs
\cite{Sinitsin2020Editable,zhu2020modifying,decao2021editing}.
Here we introduce a different setting where we have access to
new unlabeled text after model training, which must be used implicitly to update the factual predictions of the model.
In this case the update procedure also needs to figure out \textit{which}
facts must be updated and which ones remain the same.

\citet{petroni2019language} introduced the LAMA benchmark
for probing the factual knowledge memorized by LMs,
which consists of cloze queries about facts,
e.g. ``Dante was born in \blank''.
Follow up studies have introduced improved prompts
for eliciting such knowledge \cite{jiang-etal-2020-know}
as well as multilingual versions 
\cite{jiang-etal-2020-x,kassner-etal-2021-multilingual}.
However, all these benchmarks assume a static view of the knowledge
inside an LM, and consider all answers across time to be
correct for a given query.
The \templama dataset instead focuses on relations
where the answers change with time
and uses temporal scopes
to determine the correct answer.

\templama is similar in spirit to KB-QA benchmarks which focus
on temporal reasoning such as TempQuestions \cite{jia2018tempquestions}
and CronQuestions \cite{saxena-etal-2021-temporal}.
Its format, however, mimics the masked LM task typically used in pretraining,
since it is intended as a zero/few-shot probe.
Unlike those datasets, we further restrict the queries to subject and
relation pairs for which multiple
objects exist at different points in time,
and ensure a balanced distribution over the entire time period of
interest from $2010$-$2020$.

\section{Conclusion}

Though temporally-scoped facts are common in practice,
there has been little prior work exploring how these are encoded
in pretrained LMs.
We show that T5 does poorly on such facts
and training on the news domain improves it significantly.
However, simply training on more data is sub-optimal;
conditioning on the temporal context of the data improves
memorization of facts further.
Hence, we propose a time-aware language model
which conditions on string prefixes of time.
Other benefits of time-aware LMs include a better
calibration of expected changes in the future,
and a cheaper adaptation to new slices of timestamped data.

\section*{Acknowledgements}
We would like to thank the Action Editor and Reviewers
for comments on an earlier draft of this work,
and the T5X team at Google for their T5 implementation.

\section*{Supplementary Material}
\appendix

\section{\templama Templates}
\label{app:templama}

\autoref{tab:templama-templates} lists the $9$ WikiData relations used for constructing
\templama.
We instantiate the template for the relation in each fact by replacing
``\textless{}subject\textgreater'' with the name of the subject entity,
and ``\textless{}object\textgreater'' with ``\blank''.
The answer to the query is the name of the corresponding object entity.
We construct a separate query for each year that the fact is valid.

\section{Future Relations}
\label{app:future-relations}

\autoref{tab:future-relations} shows the queries used as part of
the Future Relations experiment in \S~\ref{sec:exp-calibration}.
These queries were constructed by searching for lists of events
,
popular athletes
, 
and issuing targeted queries to the WikiData Query Service.

\begin{landscape}
\begin{table}[p]
\centering
\scriptsize
\begin{tabular}{@{}lll@{}}
\toprule
\textbf{Frequent}                                                                    & \textbf{Rare}                                                              & \textbf{Never}                                                      \\ \midrule
\multicolumn{3}{c}{\textbf{Cities}}                                                                                                                                                                          \\ \midrule
The Super Bowl will take place in \_X\_.                                    & Visa Inc.'s headquarters are located in \_X\_.                    & South by Southwest will take place in \_X\_.               \\
The NCAA Men's Final Four will take place in \_X\_.                         & SEGA of America's headquarters are located in \_X\_.              & Lollapalooza will take place in \_X\_.                     \\
The first game of the World Series will take place in \_X\_.                & Barack Obama lives in \_X\_.                                      & Summerfest will take place in \_X\_.                       \\
The US PGA Championship will take place in \_X\_.                           & Hillary Clinton lives in \_X\_.                                   & Outside Lands will take place in \_X\_.                    \\
The golf US Open will take place in \_X\_.                                  & Donald Trump works in \_X\_.                                      & Spoleto Festival USA will take place in \_X\_.             \\
The NBA all-star game will take place in \_X\_.                             & The Chargers play their home games in \_X\_.                      & CMA Music Festival will take place in \_X\_.               \\
The NFL Draft will take place in \_X\_.                                     & The Raiders play their home games in \_X\_.                       & Made in America Festival will take place in \_X\_.         \\
The Netroots Nation conference will take place in \_X\_.                    & The Rams play their home games in \_X\_.                          & The US Open Tennis Championships will take place in \_X\_. \\
The MLB all-star game will take place in \_X\_.                             & General Electric's headquarters are located in \_X\_.             & The Masters tournament will take place in \_X\_.           \\
The team from \_X\_ won the NBA championship.                               & Toyota's US headquarters are located in \_X\_.                    & The Kentucky Derby will take place in \_X\_.               \\
The team from \_X\_ won the Stanley Cup.                                    & Nestle's headquarters are located in \_X\_.                       & The capital of Washington state is \_X\_.                  \\
The team from \_X\_ won the World Series.                                   & Tesla's headquarters are located in \_X\_.                        & The capital of California state is \_X\_.                  \\
The team from \_X\_ won the Super Bowl.                                     & Lebron James plays in \_X\_.                                      & The capital of Texas is \_X\_.                             \\
The golf US Women's Open will take place in \_X\_.                          & Tom Brady plays in \_X\_.                                         & The capital of Florida is \_X\_.                           \\
Wrestlemania will take place in \_X\_.                                      & Kevin Durant plays in \_X\_.                                      & The Space Needle is located in \_X\_.                      \\
                                                                            & Stephen Curry plays in \_X\_.                                     & The Statue of Liberty is located in \_X\_.                 \\
                                                                            & Sidney Crosby plays in \_X\_.                                     & Golden Gate Bridge is located in \_X\_.                    \\
                                                                            & Mike Trout plays in \_X\_.                                        & The White House is located in \_X\_.                       \\
                                                                            & The Democratic National Convention will next take place in \_X\_. & The Liberty Bell is located in \_X\_.                      \\
                                                                            & The Republican National Convention will next take place in \_X\_. &                                                            \\ \midrule
\multicolumn{3}{c}{\textbf{Countries}}                                                                                                                                                                       \\ \midrule
The Six Nations Championship will be held in \_X\_.                         & The UN Secretary general is from \_X\_.                           & The Oxford Literary Festival will take place in \_X\_.     \\
The Association for Computational Linguistics will meet in \_X\_.           & The Pope hails from \_X\_.                                        & Wimbledon will take place in \_X\_.                        \\
The Neural Information Processing Systems conference will be held in \_X\_. & The FIFA world cup was lest held in \_X\_.                        & Tomorrowland will take place in \_X\_.                     \\
The Palme d'Or winner is from \_X\_.                                        & The Cricket world cup was last held in \_X\_.                     & Hajj will take place in \_X\_.                             \\
The Tour De France winner is from \_X\_.                                    & The UEFA European Football Championship was last held in \_X\_.   & The Eiffel Tower is located in \_X\_.                      \\
The Wimbledon Men's Singles winner is from \_X\_.                           & The Olympics were last held in \_X\_.                             & The Taj Mahal is located in \_X\_.                         \\
The UEFA Champions League final will take place in \_X\_.                   & The Winter Olympics were last held in \_X\_.                      & Burj Khalifa is located in \_X\_.                          \\
The G20 summit will be held in \_X\_.                                       & The FIFA world cup was last won by \_X\_.                         & Machu Picchu is located in \_X\_.                          \\
The G7 summit will be held in \_X\_.                                        & The Cricket world cup was last won by \_X\_.                      & Stonehenge is located in \_X\_.                            \\
The United Nations Climate Change conference will take place in \_X\_.      & \_X\_ won the most gold medals in the last Olympics.              & The world's largest country by land area is \_X\_.         \\
                                                                            &                                                                   & The world's longest river is in \_X\_.                     \\
                                                                            &                                                                   & The world's tallest mountain is in \_X\_.                  \\ \bottomrule
\end{tabular}
\caption{The Future Relations dataset used to test model calibration over future years.
The three columns represent queries whose answers, intuitively, change \textit{frequently} or
every year,
\textit{rarely} or once every few years,
and \textit{never}.
The top section includes queries whose answer is a US city,
while the bottom section includes queries whose answer is a country.
}
\label{tab:future-relations}
\end{table}
\end{landscape}

\bibliography{anthology,custom}
\bibliographystyle{acl_natbib}

\end{document}